\pdfoutput=1

\documentclass[11pt]{article}

\usepackage[]{acl}
\usepackage{graphicx}
\usepackage{tabularx} 

\usepackage{multirow}   
\usepackage{xcolor}     
\usepackage{colortbl}
\usepackage{amsmath,amsfonts}
\usepackage{algorithmic}
\usepackage{adjustbox}
\usepackage{algorithm}
\usepackage{array}
\usepackage{subfigure}
\usepackage{subcaption}
\usepackage{textcomp}
\usepackage{stfloats}
\usepackage{url}
\usepackage{verbatim}
\usepackage{graphicx}
\usepackage{booktabs}
\usepackage{graphicx}%
\usepackage{float}
\usepackage{caption}
\usepackage{times}
\usepackage{latexsym}
\usepackage{natbib}
\usepackage{balance}
\usepackage[normalem]{ulem}

\definecolor{lightgreen}{RGB}{173, 216, 230}
\definecolor{lightred}{RGB}{255, 218, 185}
\definecolor{lightpurple}{RGB}{216, 191, 216} 
\definecolor{lightpinkred}{RGB}{255, 182, 193} 

\usepackage{makecell}
\usepackage{caption}
\usepackage{xurl}
\usepackage{amsmath}
\usepackage[table]{xcolor}  
\usepackage{soul}

\usepackage{array}
\usepackage{geometry}

\usepackage{tikz}
\usepackage{enumitem}
\usepackage{arydshln}

\usepackage[T1]{fontenc}

\usepackage[utf8]{inputenc}

\usepackage{microtype}

\usepackage{inconsolata}

\usepackage{graphicx}

\def\method{Genii}

\title{Mitigating Judgment Preference Bias in Large Language Models through Group-Based Polling}

\author{Shuliang Liu$^{1}$\thanks{ \ \ indicates equal contribution.}, Zhipeng Xu$^{1}$\footnotemark[1], Zhenghao Liu$^{1}$\thanks{ \ \ indicates corresponding author.}, Yukun Yan$^{2}$,\\ 
\textbf{Minghe Yu$^{3}$, Yu Gu$^{1}$, Chong Chen$^{4}$, Huiyuan Xie$^{2}$ and Ge Yu$^{1}$} \\ 
$^1$School of Computer Science and Engineering, Northeastern University, Shenyang, China \\
$^2$Department of Computer Science and Technology, Tsinghua University, Beijing, China \\
$^3$Software College, Northeastern University, Shenyang, China \\
$^4$Huawei, China\\
}

\begin{document}
\maketitle

\begin{abstract}
Large Language Models (LLMs) as automatic evaluators, commonly referred to as ``LLM-as-a-Judge'', have also attracted growing attention. This approach plays a vital role in aligning LLMs with human judgments, providing accurate and reliable assessments. 
However, LLM-based judgment models often exhibit judgment preference bias during the evaluation phase, tending to favor responses generated by themselves, undermining the reliability of their judgments.
This paper introduces the \textbf{G}roup-Bas\textbf{e}d Polli\textbf{n}g Opt\textbf{i}m\textbf{i}zation (\method{}), an unsupervised multi-agent collaborative optimization framework that mitigates the inherent judgment preference bias of judgment models.
Specifically, \method{} integrates various LLM-based judgment models into a multi-agent system and simulates the interactive client-server polling mechanism to optimize each client agent unsupervisedly. Our experiments demonstrate that \method{} outperforms supervised models trained on annotated judgment data, while requiring no human-labeled annotations.
\method{} consistently improves performance across different client agents during the polling, even when weaker models act as server agents.
Further analysis reveals that \method{} effectively mitigates judgment preference bias of LLM-based
judgment models, demonstrating its effectiveness. All codes are available at \url{https://github.com/NEUIR/Genii}.

\end{abstract}

\section{Introduction}

Large Language Models (LLMs), such as GPT-4~\cite{openai2023gpt} and Gemini~\cite{team2023gemini}, have demonstrated impressive capabilities across a wide range of downstream Natural Language Processing (NLP) tasks~\cite{zhao2023survey,wei2022emergent,hou2024large,huang2025pip,yao2025llm}.
Beyond positing LLMs as answer generators, recent research has explored using LLMs as evaluators to automatically assess response quality of models, a paradigm known as LLM-as-a-Judge~\cite{zheng2023judging,gu2024survey,tan2024judgebench,li2025generation}. 
This approach offers a scalable and cost-effective alternative to human annotation, demonstrating promising alignment with human judgments and has been widely used in the assessment of response quality in language models~\cite{kim2023prometheus,wu2024meta,yang2024crag}. 
Despite their notable success, LLM-based evaluators often face challenges in judgment preference bias--a systematic tendency of LLMs to favor responses generated by themselves or aligned with their generation patterns~\cite{bitton2023visit,wataoka2024self,gu2024survey, chen2025llm}, undermining the reliability of their judgments~\cite{koo2023benchmarking}.

\begin{figure}[t]
    \centering
    \includegraphics[width=\columnwidth]{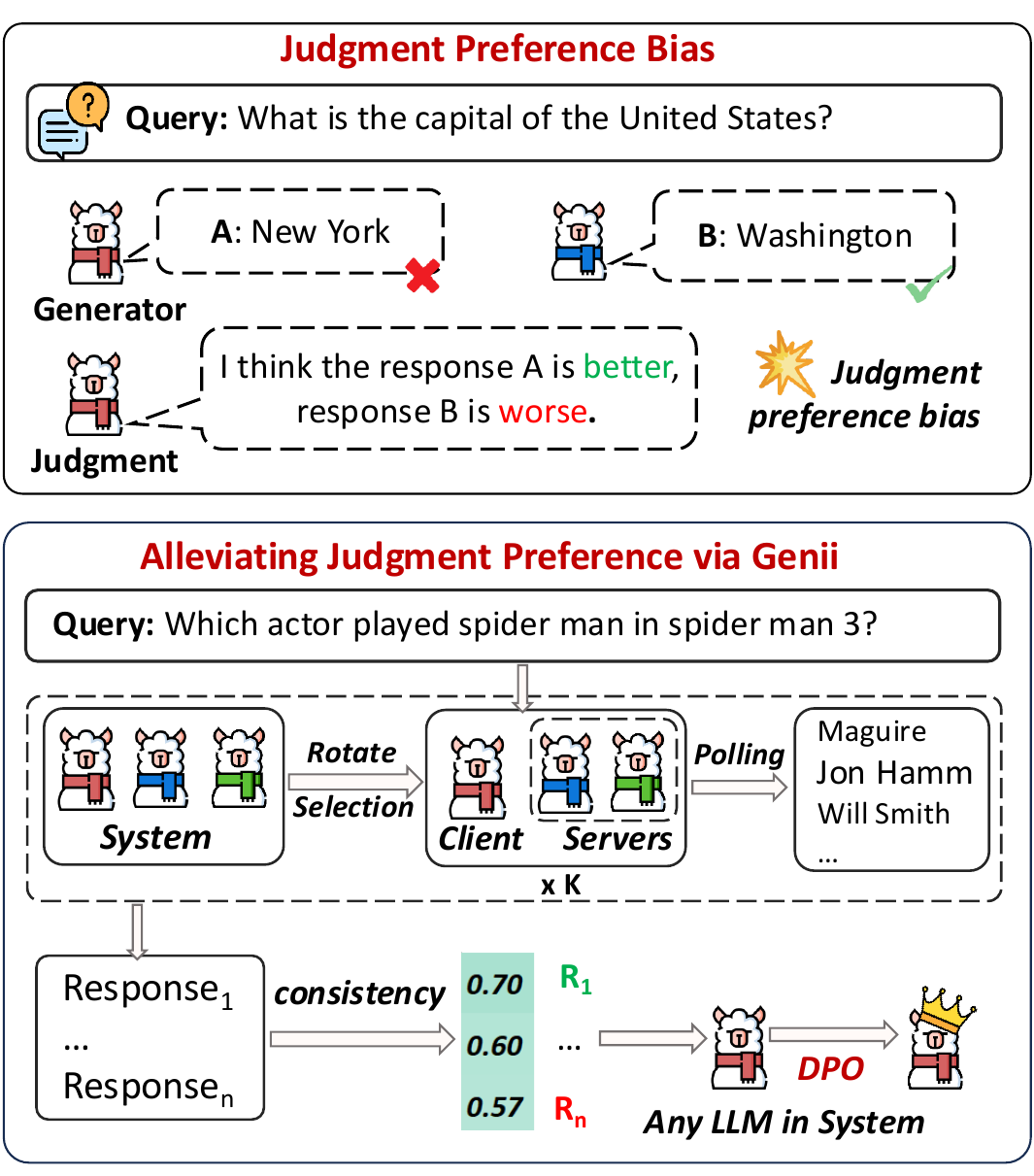}
    \caption{The Framework of \method{}. It alleviates judgment preference of LLMs during the evaluation process.}
    \label{fig:example}
\end{figure}
To mitigate judgment preference bias of LLM-based evaluators and provide more accurate assessments, researchers have made extensive efforts~\cite{gu2024survey,wang2024dhp,he2023socreval}. 
Previous studies have focused on enhancing the reasoning capabilities of LLM-based evaluators to generate more rational judgments~\cite{yang2025rethinking}. 
Specifically, they prompt LLMs to generate chain-of-thought reasoning before making judgments~\cite{wei2022emergent,gallegos2024self}, engage in self-consistency decoding~\cite{wang2022self}, or apply self-debiasing to correct potentially biased decisions~\cite{ganguli2023capacity}. 
Nevertheless, LLMs often still overlook their own biased behavior~\cite{wang2022self}, avoiding key comparison questions and giving vague statements that fail to identify or correct their underlying preference bias~\cite{yang2025rethinking}.
Recent work has shifted away from relying on a single evaluation model, instead harnessing collaboration among multiple models to obtain more objective and reliable judgments~\cite{park2024offsetbias,liu2025assistant}. 
While such ensemble-based methods improve robustness, they often rely on stronger models to guide weaker ones, and are only applied at the inference stage.
However, allowing knowledge to flow in both directions during optimization, so that weaker models can help improve stronger models without hurting their performance, remains a major challenge~\cite{zhang2018deep}.


In this paper, we propose \textbf{G}roup-Bas\textbf{e}d Polli\textbf{n}g Opt\textbf{i}m\textbf{i}zation (\method{}), an unsupervised multi-agent collaborative optimization framework that encourages multiple LLM judgment models to interact with each other to mitigate their own inherent judgment preference bias.
As illustrated in Figure~\ref{fig:example}, instead of introducing additional preference datasets, \method{} integrates various LLM-based judgment models into a multi-agent system and simulates the interactive client-server polling mechanism to optimize each agent unsupervisedly.
\method{} first selects one model in the group as the client agent, responsible for initiating the polling request. 
The remaining models act as server agents to evaluate and provide feedback on every client request. 
Then \method{} computes the group consistency score for each request. By rotating the client agent to collect responses from all agents, \method{} aggregates the group consistency scores to curate preference pairs, thereby optimizing each agent towards the collective preference of the group and mitigating individual preference bias.

Our experimental results demonstrate that \method{} outperforms supervised models trained on annotated judgment data, without requiring human-labeled annotations, achieving a significant improvement over vanilla LLMs. 
\method{} consistently improves model performance regardless of which model serves as the client agent during the polling. 
Remarkably, even when weaker models act as server agents, their feedback contributes meaningfully to the optimization of stronger models. 
Further analysis shows that with the improvement of LLMs’ generation capabilities, their judgment performance also enhances, demonstrating the effectiveness of \method{}.
Additionally, \method{} effectively mitigates the preference bias of LLM-based judgment models toward their incorrect answers and enables them to yield more reliable judgments.

\section{Related Work}
Large Language Models (LLMs) as automatic evaluators, or ``LLM-as-a-Judge'', have proven effective across various tasks~\cite{zhang2024self,zhang2024balancing}. By prompting LLMs to decide which response is better according to specific evaluation criteria, they can produce judgments closely aligned with human preferences~\cite{lin2023llm}. Most existing LLM-based judgment models use closed-source APIs such as GPT-4 to score or rank responses and generate explanations~\cite{zheng2023judging}, but these APIs introduce significant cost and reduce reproducibility~\cite{gu2024survey}. Recent work further fine-tunes LLMs for judgment, showing that with suitable training, LLMs can be highly reliable evaluators~\cite{zhu2023judgelm,kim2023prometheus,kim2024prometheus2}.
Subsequent work explores self-improvement approaches that enhance LLMs' judgment ability without relying on additional supervision to reduce cost~\cite{liu2025consjudge}.
However, the reliability of LLMs as evaluators remains unclear, as they may be affected by judgment preference bias~\cite{zheng2023judging}.

To improve the reliability of LLM-based judgment models, it is crucial to explore effective strategies for mitigating judgment preference bias.
Some approaches focus on strengthening the reasoning abilities of these models to yield more rational evaluations~\cite{yang2025rethinking}. For example, ~\citet{gallegos2024self} propose a self-debiasing method that prompts the model to perform two rounds of judgment, with the second aimed at correcting possible biases in the first. Self-Consistency~\cite{wang2022self} reduces bias by increasing the number of reasoning attempts and aggregating results via majority voting, while Long Reasoning~\cite{chen2025llm} extends the reasoning chain.
Both methods have demonstrated great potential to reduce bias during the evaluation phase.
However, prompt-based self-debiasing methods often lead judgment models to overlook their actual biased behaviors, and increasing the number or length of reasoning steps incurs higher computational cost~\cite{yang2025rethinking,hassid2025don}.

Moreover, recent advances have shifted evaluation from single-model paradigms toward multi-model collaborative judgment frameworks to achieve more objective and reliable evaluations.
Debate-based approaches utilize multiple LLMs to argue for opposing answers, with a superior LLM reviewing the discussion and deciding which answer is better~\cite{khan2024debating}. 
Ensemble methods further reduce preference bias by averaging scores across multiple judgment models or down-weighting cases where the individual judgment model shows unusually high confidence~\cite{wataoka2024self,chen2025beyond}.
Although these approaches help mitigate preference bias by relying on stronger models to guide weaker ones, they are typically only applied at the inference stage and tend to overlook the potential complementary value that weaker models can provide. In contrast, \method{} enables bi-directional knowledge transfer during optimization, allowing stronger models to benefit from feedback provided by weaker models, and thus yielding more accurate judgments.

\section{Methodology}
In this section, we introduce our \textbf{G}roup-Bas\textbf{e}d Polli\textbf{n}g Opt\textbf{i}m\textbf{i}zation (\method{}) framework.
We first introduce the preliminaries of LLM-based judgement models (Sec.~\ref{method:preliminary}). 
Then we describe the detailed mechanism of \method{}, which optimizes each model in the group through polling interactions (Sec.~\ref{method:method}).


\begin{figure*}[t]
\centering
\includegraphics[width=\linewidth]{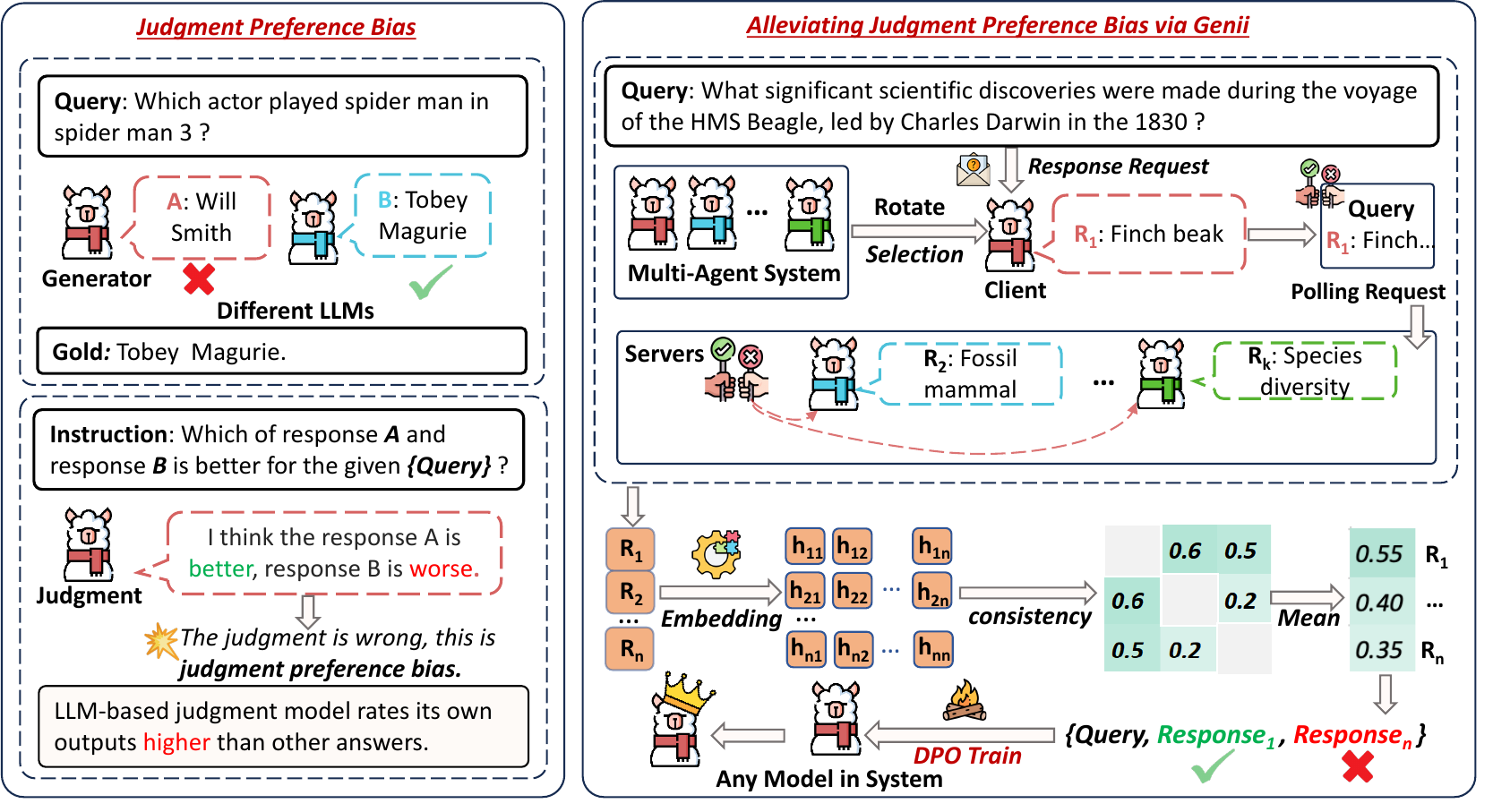}
\caption{The Framework of Our \method{} Method.} 
\label{fig:model}
\end{figure*}

\subsection{Preliminary of LLM-as-a-Judge}
\label{method:preliminary}
Given a query $q$ and two candidate answers $c_1$ and $c_2$, the goal of LLM-as-a-Judge is to identify the higher-quality answer to the query from the two candidates.

LLM-based judgment model, guided by a predefined evaluation prompt $p$, compares the quality of two candidate answers $c_1$ and $c_2$. The model produces a judgment process $e$ as well as a final judgment result $\epsilon$. This assessment process can be formulated as:
\begin{equation}\small
    (e,\epsilon) = \text{LLM}(p,q,c_1,c_2),
\end{equation}
where $\epsilon$ represents one of three possible results: $c_1$ is better, $c_2$ is better, or the two answers are of comparable quality (tie).

Current approaches for optimizing LLM-based judgment models primarily rely on leveraging guidance from stronger judgment models to improve weaker ones. However, they often fail to fully exploit the potential complementary contributions of the weaker models, which could also help further enhance the judgment performance of stronger models.

\subsection{Optimizing LLM-Based Judgment Models through Group-Based Polling}
\label{method:method} 
In this subsection, we introduce the \textbf{G}roup-Bas\textbf{e}d Polli\textbf{n}g Opt\textbf{i}m\textbf{i}zation (\method{}) method, which integrates multiple LLM-based judgment models into a multi-agent system and simulates interactive client-server polling to optimize every agent. 

\textbf{Client-Server Agent System. }
To optimize the LLM-based judgment models, \method{} first instantiates a collaborative multi-agent system composed of $m$ heterogeneous agents, which can be formulated as:
\begin{equation}\small
\mathcal{V} = \{V_1, V_2 , \ldots, V_m\},
\end{equation}
where each agent is implemented with a different LLM-based judgment model, which has a unique inherent judgment preference bias. 
Before the agents start communicating, one of the agents $V_i \in \mathcal{V}$ is configured as the client, which generates responses and initiates the polling process during the communication. While the remaining agents $V_j \in \mathcal{V}$ ($j \ne i$) act as servers to process the polling request, evaluate the response from the client independently, and provide feedback based on their own criteria. 
Then \method{} computes the group consistency score for each request. By rotating the client agent to collect responses from all agents, \method{} aggregates the group consistency scores to curate preference pairs for optimizing each agent in the system.

\textbf{Polling Sampling.} 
Once the client agent $V_i$ is configured, it receives a query $q$ and engages in response–request interactions repeatedly. 
For the $k$-th interaction, the client agent first generates a response $y_k^{(i)}$ based on $q$, which can be represented as:
\begin{equation}\small
y_k^{(i)} = V_i(q),
\end{equation}
where the original query $q$ and the response $y_k^{(i)}$ generated by the client agent constitute a polling request $r_k$. We can formulate the content of the polling request $r_k$ as:
\begin{equation}\small
r_k^{(i)} = \{y_k^{(i)}, q\},
\end{equation}
which is subsequently broadcast to the remaining server agents in the system. Each server agent $V_j \in \mathcal{V}$ ($j \ne i$) conducts a response–feedback operation: upon receiving the polling request $r_k^{(i)}$, it generates a response for the query based on its own criteria, and calculates a client-server consistency score as feedback, which can be formulated as:
\begin{equation}\small
s(i,k) = \cos\left( \text{Emb}\left(y_k^{(i)}\right),\ \text{Emb}\left(V_j(q)\right) \right),
\end{equation}
where the $\mathrm{Emb}$ denotes a text encoder that can map the response into a semantic embedding space. 
To assess the group consistency between the client and the group, for each client agent response $y_k^{(i)}$, we integrate the consistency scores from all the server agents in the group and compute the group-based consistency as:
\begin{equation}\small
S(i,k) = \frac{1}{|\mathcal{V}| - 1} \sum_{j \ne i} s(i,k),
\end{equation}
where the score reflects how well the client's response $y_k^{(i)}$ aligns with the implicit preferences of the server agents in the system. 

\textbf{Group-based Consistency Optimization.}
For the query $q$, the client agent repeats the response–request operation until the maximum number of sampling $K$ is reached. By rotating the client role across the group, we can obtain a set of responses $\mathcal{Y}$ generated by all agents, each associated with a group-based consistency score.
Based on these group-based consistency scores, we select the highest- and lowest-consistency responses from all responses $\mathcal{Y}$ to form preference pairs $(y^+, y^-)$. These pairs are then used to train each agent in the system via Direct Preference Optimization (DPO)~\cite{Rafailov2023DPO}:
\begin{equation}\small
L = -\log \sigma\left( 
\beta \log \frac{V_i(y^+ \mid q)}{V_i^{\text{ref}}(y^+ \mid q)} 
- \beta \log \frac{V_i(y^- \mid q)}{V_i^{\text{ref}}(y^- \mid q)} 
\right),
\end{equation}
where $V_i$ denotes the agent model under optimization, $V_i^{\text{ref}}$ denotes the reference model and $\sigma$ is the sigmoid function.
All models benefit from diverse group feedback and gradually converge toward more consistent and unbiased evaluations.

\section{Experimental Methodology}
This section describes the datasets, evaluation metrics, and baselines used in our experiments. More implementation details are shown in Appendix~\ref{more_details}.

\textbf{Datasets.} We describe the datasets used by \method{} for training and evaluation.

\textit{Training.} \method{} uses instructions from instruction-following datasets and a small amount of QA data to optimize the multi-agent system.
Specifically, we sample instructions from Evol-Instruct~\cite{xu2024wizardlm} and collect QA pairs from several widely used QA datasets.
In total, we construct a dataset comprising 7,992 unique samples for unsupervised training.
Additionally, we randomly sample 500 instances from UltraFeedback~\cite{cui2023ultrafeedback} to construct the development dataset.
More detailed statistics of the training data are provided in Appendix~\ref{more_details}.

\textit{Evaluation.} To assess the performance of \method{}, we select multiple tasks from prior work~\cite{li2024ddr,liu2025assistant}, including accuracy evaluation on UltraFeedback, MTBench~\cite{zheng2023judging}, AutoJ~\cite{li2023autoj}, PreferenceBench~\cite{kim2024prometheus2}, and RewardBench~\cite{lambert2024rewardbench}, as well as judgment preference bias evaluation on NQ~\cite{Kwiatkowski2019nq}, HotpotQA~\cite{Yang2018hotpotqa}, and TriviaQA~\cite{Joshi2017triviaqa}.
For NQ, HotpotQA, and TriviaQA, we build the preference label following the prior work~\cite{chen2025llm}.
The detailed data statistics are shown in Table~\ref{table1:testdataset}.

\textbf{Evaluation Metrics.} 
We employ accuracy and Harmful Self-Preference Propensity (HSPP) as evaluation metrics to evaluate the judgment performance of LLM-based judgment models. The prompt templates
used for evaluation are shown in Appendix~\ref{prompt}.

\textit{Accuracy.} We follow~\citet{Mariu2019pairwise} and use accuracy as the evaluation metric. In judgment benchmarks, each example consists of a query $q$ along with two annotated candidate responses, $y_1$ and $y_2$, corresponding to the chosen and rejected answers. We denote $\mathcal{Y}^* = \{y^*_1, y^*_2, \ldots, y^*_n\}$ as the set of all chosen responses. For each example, the judgment model selects the better answer between the two candidates. Let $\mathcal{R} = \{r_1, r_2, \ldots, r_n\}$ represents the set of better answers. Finally, we can compute the judgment accuracy score based on these judgment results:
\begin{equation}\small
\mathrm{Acc} = \frac{1}{n} \sum_{k=1}^n \mathbb{I}\left[ r_k = y_k^* \right],
\end{equation}
where $\mathbb{I}[\cdot]$ is an indicator function that evaluates to 1 if the judgment result $r_{k}$ matches the chosen response $y_{k}^*$, and 0 otherwise.


\begin{table}[t]
\centering
\small
\begin{tabular}{llr}
\toprule
\textbf{Dataset} & \textbf{Metric} & \textbf{Total} \\
\midrule
UltraFeedback~\shortcite{cui2023ultrafeedback} & Accuracy & 2,000 \\
MTBench~\shortcite{zheng2023judging} & Accuracy & 3,355 \\
AutoJ~\shortcite{li2023autoj} & Accuracy & 1,392 \\
PreferenceBench~\shortcite{kim2024prometheus2} & Accuracy & 1,998 \\
RewardBench~\shortcite{lambert2024rewardbench} & Accuracy & 2,985 \\
\midrule
NQ~\shortcite{Kwiatkowski2019nq} & HSPP & 2,837 \\
TriviaQA~\shortcite{Joshi2017triviaqa} & HSPP & 5,359 \\
HotpotQA~\shortcite{Yang2018hotpotqa} & HSPP & 5,600 \\
\bottomrule
\end{tabular}
\caption{Data Statistics of the Evaluation Datasets.}
\label{table1:testdataset}
\vspace{-1em}
\end{table}

\textit{Harmful Self-Preference Propensity.} 
We follow~\citet{chen2025llm} to compute HSPP to quantify judgment preference bias. 
Let $\mathcal{D}_{\text{diff}}$ denote a specially constructed evaluation set, where each instance contains two distinct responses, $y'_i$ and $y'_j$, produced by two different vanilla judgment models $\mathcal{M}_i$ and $\mathcal{M}_j$ for the same question $q$.
In each instance, exactly one of $y'_i$ or $y'_j$ is correct, but not both, and $y^*$ denotes the correct one.
We can then compute HSPP for model $\mathcal{M}_i$ as follows:



\begin{equation}\small
\begin{aligned}
\mathrm{HSPP}_{\mathcal{M}_i}
&=
\frac{
\sum_{\mathcal{M}_j \in \mathcal{G}^*}
\sum_{q \in \mathcal{D}_{\mathrm{diff}}}
\phi(q)
}{
\sum_{\mathcal{M}_j \in \mathcal{G}^*}
\sum_{q \in \mathcal{D}_{\mathrm{diff}}}
\mathbb{I}\{\,y^* = y'_j\,\}
}, \\[3pt]
\phi(q)
&= \mathbb{I}\{\,y'_i = \mathcal{M}_i(q, y'_i, y'_j), y^*=y'_j\},
\end{aligned}
\end{equation}
where $\mathbb{I}\{\cdot\}$ is an indicator function that returns 1 if both conditions are satisfied, and 0 otherwise. $\mathcal{G}^*$ represents the agent system excluding model $\mathcal{M}_i$. A higher HSPP indicates that the judgment model exhibits a greater tendency toward preference bias.

\textbf{Baselines.} In our experiments, we compare \method{} with five judgment models. The Vanilla LLM method directly uses the LLM as the judgment model and then leverages the evaluation prompts to ask it to produce the judgments. The Reprompt method prompts the LLM to correct potentially biased decisions in its initial judgment to reduce bias~\cite{gallegos2024self}. 
The Self-Consistency method generates evaluation results multiple times and aggregates them via majority voting~\cite{wang2022self}.
The Long Reasoning method directly employs Reasoning Language Models (RLMs) distilled from DeepSeek-R1~\cite{guo2025deepseek} as the judgment models~\cite{chen2025llm}.
The Judgment SFT method further fine-tunes LLMs using annotation data generated by Qwen3-32B~\cite{Bai2023qwen} on the UltraFeedback dataset, which has been used in previous work~\cite{kim2023prometheus} to improve the judgment performance of LLMs.

\section{Evaluation Results}
In this section, we first show the performance of various judgment models. Then, we conduct ablation studies to examine the impact of different modules, and further conduct a cost–performance analysis to demonstrate the scalability of \method{}. Subsequently, we evaluate the effectiveness of \method{} in optimizing LLMs for evaluation. More experiments are shown in Appendix~\ref{app:judgment_consistency}-~\ref{position_bias}. We also conduct several case studies in Appendix~\ref{case}.

\begin{table*}[ht]
\centering
\small
\resizebox{\textwidth}{!}{
\begin{tabular}{lcccccccc}
\toprule
\multirow{2}{*}{\textbf{Method}} & \textbf{UltraFb} & \textbf{MT} & \textbf{AutoJ} & \textbf{Prefer} & \textbf{Reward} & \textbf{NQ} & \textbf{HotpotQA} & \textbf{TriviaQA} \\
& {(Acc)$\uparrow$} & {(Acc)$\uparrow$} & {(Acc)$\uparrow$} & {(Acc)$\uparrow$} & {(Acc)$\uparrow$} & {(HSPP)$\downarrow$} & {(HSPP)$\downarrow$} & {(HSPP)$\downarrow$} \\
\midrule
\multicolumn{9}{c}{\cellcolor{gray!8}\textbf{Different Model Families}} \\
\midrule
\multicolumn{9}{l}{\textit{Llama3.1-8B-Instruct}} \\
\midrule
Vanilla & 47.20 & 40.86 & 38.86 & 71.07 & 50.28 & 6.96 & 3.09 & 3.95 \\
Reprompt & 35.10 & 31.43 & 27.30 & 56.36 & 40.20 & 40.51 & 37.89 & 38.16 \\
Self-Consistency & 47.70 & 42.40 & 39.35 & 75.23 & 50.50 & 6.33 & 2.84 & \textbf{1.32} \\
Long Reasoning & 55.30 & \textbf{51.27} & \textbf{50.43} & 78.53 & 66.30 & \textbf{1.00} & 3.28 & \textbf{1.32} \\
Judgment SFT & \textbf{56.70} & 46.26 & 48.99 & \textbf{81.68} & \textbf{71.06} & 5.70 & 3.09 & 2.64 \\
\textbf{\method{}} & 45.95 & 43.40 & 40.73 & 66.17 & 51.15 & 5.06 & \textbf{1.55} & \textbf{1.32} \\
\midrule
\multicolumn{9}{l}{\textit{Gemma2-9B-Instruct}} \\
\midrule
Vanilla & 59.90 & 52.16 & 47.05 & 78.48 & 64.66 & 32.37 & 40.61 & 46.44 \\
Reprompt & 54.10 & 44.62 & 44.58 & 70.12 & 60.40 & 39.93 & 39.70 & 46.10 \\
Self-Consistency & 61.54 & 53.00 & 47.88 & 80.48 & 66.36 & 38.85 & 44.51 & 48.14 \\
Judgment SFT & 59.60 & 49.54 & 50.43 & 79.83 & \textbf{73.77} & 30.10 & 39.16 & 44.07 \\
\textbf{\method{}} & \textbf{63.45} & \textbf{56.13} & \textbf{51.87} & \textbf{82.13} & 67.17 & \textbf{28.90} & \textbf{34.39} & \textbf{40.68} \\
\midrule
\multicolumn{9}{l}{\textit{Qwen2.5-7B-Instruct}} \\
\midrule
Vanilla & 62.25 & 55.44 & 48.35 & 82.43 & 69.41 & 63.49 & 62.93 & 70.13 \\
Reprompt & 58.00 & 50.31 & 46.41 & 70.82 & 66.60 & 61.22 & 63.58 & 68.87 \\
Self-Consistency & 62.15 & 55.12 & 48.88 & 83.00 & 69.57 & 65.20 & 67.24 & 74.96 \\
Long Reasoning & 54.65 & 48.26 & 45.40 & 74.02 & 63.95 & 62.30 & 60.96 & 64.25 \\
Judgment SFT & 57.25 & 50.52 & 51.15 & 78.58 & 71.48 & 58.66 & 53.23 & 51.88 \\
\textbf{\method{}} & \textbf{64.75} & \textbf{56.42} & \textbf{53.38} & \textbf{85.34} & \textbf{73.60} & \textbf{52.13} & \textbf{50.43} & \textbf{49.19} \\
\midrule
\multicolumn{9}{c}{\cellcolor{gray!8}\textbf{Same Model Family}} \\
\midrule
\multicolumn{9}{l}{\textit{Qwen2.5-3B-Instruct}} \\
\midrule
Vanilla & 55.30 & 45.66 & 41.59 & \textbf{66.07} & \textbf{57.89} & 35.08 & 40.52 & 31.98 \\
\textbf{\method{}} & \textbf{55.95} & \textbf{49.51} & \textbf{42.03} & 61.96 & 56.65 & \textbf{17.33} & \textbf{21.12} & \textbf{17.35} \\
\midrule
\multicolumn{9}{l}{\textit{Qwen2.5-7B-Instruct}} \\
\midrule
Vanilla & 62.25 & \textbf{55.44} & 48.35 & \textbf{82.43} & 69.41 & 63.49 & 62.93 & 70.13 \\
\textbf{\method{}} & \textbf{63.15} & 54.96 & \textbf{51.80} & 78.46 & \textbf{72.63} & \textbf{29.55} & \textbf{34.70} & \textbf{40.79} \\
\midrule
\multicolumn{9}{l}{\textit{Qwen2.5-14B-Instruct}} \\
\midrule
Vanilla & 65.30 & 56.99 & 54.60 & 86.64 & 80.80 & 48.15 & 45.47 & \textbf{21.40} \\
\textbf{\method{}} & \textbf{68.75} & \textbf{58.42} & \textbf{56.68} & \textbf{87.49} & \textbf{82.48} & \textbf{38.07} & \textbf{39.01} & 31.13 \\
\bottomrule
\end{tabular}
}
\caption{Overall Performance of Different Judgment Models. The highest scores are highlighted in \textbf{bold}. UltraFb denotes UltraFeedback, while MT, Prefer, and Reward refer to MTBench, PreferenceBench, and RewardBench.}
\label{table1:overall}
\end{table*}

\subsection{Overall Performance}
\label{sec:overall}
This section presents the overall performance of \method{} and baseline models across different tasks. 

As shown in Table~\ref{table1:overall}, we first compare \method{} with five judgment models, including Vanilla, Reprompt, Self-Consistency, Long Reasoning, and Judgment SFT. 
These models are evaluated within an agent system consisting of models from different families, including the Llama3.1, Qwen2.5, and Gemma2 backbones.
Compared to Vanilla LLMs, \method{} achieves better performance
for enhancing the judgment quality of LLM-based judgment models, indicating its effectiveness. 
The Reprompt method, which encourages judgment models to identify and correct potentially biased judgments, can even degrade judgment performance. 
This degradation is due to limited-capacity judgment models being unable to recognize their own biases, thereby diminishing judgment performance effectively.
In contrast, \method{} enables these weaker models to yield more accurate and reliable judgments.
Self-Consistency and Long Reasoning methods also achieve significant improvements over Vanilla LLMs, demonstrating that increasing reasoning cost can help reduce errors in assessments. 
In contrast, \method{} delivers accurate judgments with much shorter reasoning time, striking a better balance between efficiency and performance.
Additionally, compared to Judgment SFT, \method{} employs group-based consistency optimization, achieving comparable or even superior judgment accuracy and effectively mitigating judgment biases in an unsupervised approach without human-labeled annotations.

\begin{table*}[t]
\centering
\small
\resizebox{\textwidth}{!}{
\begin{tabular}{lcccccccc}
\toprule
\multirow{2}{*}{\textbf{Method}} & \textbf{UltraFb} & \textbf{MT} & \textbf{AutoJ} & \textbf{Prefer} & \textbf{Reward} & \textbf{NQ} & \textbf{HotpotQA} & \textbf{TriviaQA} \\
& {(Acc)$\uparrow$} & {(Acc)$\uparrow$} & {(Acc)$\uparrow$} & {(Acc)$\uparrow$} & {(Acc)$\uparrow$} & {(HSPP)$\downarrow$} & {(HSPP)$\downarrow$} & {(HSPP)$\downarrow$} \\
\midrule
\multicolumn{9}{l}{\textit{Llama3.1-8B-Instruct}} \\
\midrule
\textbf{\method{} ($K=5$)} & \textbf{45.95} & \textbf{43.40} & 40.73 & 66.17 & 51.15 & 5.06 & \textbf{1.55} & \textbf{1.32} \\
w/o DPO & 46.20 & 42.65 & 38.29 & 64.31 & 49.58 & 5.70 & 3.61 & 2.63 \\
w/o Consistency & 45.90 & 43.04 & \textbf{42.96} & \textbf{69.47} & \textbf{53.60} & 5.70 & 3.35 & 2.63 \\
w/o Gemma & 47.15 & 43.31 & 38.51 & 66.72 & 50.65 & 6.96 & 2.06 & 2.63 \\
w/o Gemma, Qwen & 44.90 & 41.04 & 38.36 & 65.22 & 46.80 & 9.49 & 3.61 & 3.96 \\
\method{} ($K=3$) & 45.60 & 43.19 & 39.66 & 66.32 & 51.59 & \textbf{3.16} & 3.35 & \textbf{1.32} \\
\midrule
\multicolumn{9}{l}{\textit{Gemma2-9B-Instruct}} \\
\midrule
\textbf{\method{} ($K=5$)} & \textbf{63.45} & \textbf{56.13} & \textbf{51.87} & \textbf{82.13} & 67.17 & \textbf{28.90} & \textbf{34.39} & \textbf{40.68} \\
w/o DPO & 59.20 & 51.48 & 46.05 & 75.93 & 61.47 & 51.32 & 56.07 & 62.71 \\
w/o Consistency & 59.05 & 49.30 & 45.83 & 75.43 & 65.03 & 40.29 & 44.94 & 51.86 \\
w/o Qwen & 62.55 & 55.35 & 48.99 & 80.13 & \textbf{70.39} & 36.09 & 39.31 & 47.12 \\
w/o Llama, Qwen & 56.10 & 47.90 & 44.90 & 67.87 & 61.91 & 36.93 & 43.35 & 50.17 \\
\method{} ($K=3$) & 61.70 & 53.98 & 51.65 & 81.23 & 66.77 & 35.61 & 36.27 & 44.07 \\
\midrule
\multicolumn{9}{l}{\textit{Qwen2.5-7B-Instruct}} \\
\midrule
\textbf{\method{} ($K=5$)} & 64.75 & \textbf{56.42} & \textbf{53.38} & \textbf{85.34} & \textbf{73.60} & 52.13 & \textbf{50.43} & \textbf{49.19} \\
w/o DPO & 56.65 & 45.78 & 46.48 & 74.12 & 67.37 & 53.69 & 54.31 & 54.56 \\
w/o Consistency & 63.75 & 54.16 & 49.43 & 78.98 & 71.02 & 53.84 & 50.86 & 57.25 \\
w/o Gemma & 64.20 & 56.24 & 53.23 & 84.78 & 73.00 & 54.97 & 51.51 & 50.98 \\
w/o Llama, Gemma & 61.80 & 52.07 & 48.35 & 72.42 & 71.39 & 52.98 & 52.16 & 56.89 \\
\method{} ($K=3$) & \textbf{64.85} & 56.18 & 52.66 & 83.98 & 70.75 & \textbf{51.12} & 51.29 & 55.46 \\
\bottomrule
\end{tabular}
}
\caption{Ablation Study. The highest scores are highlighted in \textbf{bold}. $K$ represents the maximum number of response–request sampling operations for each agent. ``w/o Gemma'' and ``w/o Gemma, Qwen'' indicate that Gemma, or both Gemma and Qwen, are removed from the agent system, respectively.}
\label{table1:ablation}
\end{table*}

We further analyze the generalization capability of \method{} within agent systems composed of models from the same model family.
Specifically, we assemble three Qwen2.5 models (3B, 7B, 14B) into a multi-agent system and jointly optimize them using \method{}. 
Experimental results show that \method{} not only leverages stronger models (7B, 14B) as server agents to enhance the weaker client agent (3B), but also significantly improves the strongest client agent (14B) when using feedback from the weakest model (3B) as the server agent, consistently enhancing the performance of all agents across all datasets.
Furthermore, the results show that the Qwen2.5-7B-Instruct model optimized with \method{} achieves comparable performance to the Vanilla Qwen2.5-14B-Instruct across multiple evaluation tasks. This finding suggests that, although the multi–agent system introduces additional computational overhead, it yields substantial performance gains, effectively allowing small judgment models to achieve comparable performance to larger judgment models.


\subsection{Ablation Study}
\label{sec:ablation}
This experiment conducts ablation studies to investigate the effectiveness of different modules and training strategies in \method{}. 

As shown in Table~\ref{table1:ablation}, we evaluate the judgment performance of six \method{} variants. Specifically, \method{} w/o DPO directly uses positive responses from the DPO training set for supervision, while \method{} w/o Consistency randomly selects chosen and rejected responses for DPO training. We also compare performance after removing one or two agents from the system, and define \method{} ($K=3$) as having each agent perform up to three response–request sampling operations per input.
Compared to \method{} w/o DPO, the standard \method{} achieves better results, since the DPO method enables the model to better distinguish between high- and low-quality responses, thus improving its judgment ability. When the consistency score is not used and chosen-rejected response pairs are selected randomly, there is a significant drop in judgment performance, demonstrating the importance of our unsupervised preference pair construction.
In addition, in both the Gemma and Qwen backbones, removing the weaker Llama model causes a clear performance drop, showing that Genii enables weaker agents to effectively enhance stronger ones.
Finally, compared with \method{} ($K=3$), \method{} ($K=5$) with more response–request operations achieves greater improvements in judgment performance, demonstrating that increasing the number of samples can further enhance preference data quality for training.

\begin{table}[t]
\centering
\small
\begin{tabular}{lccc}
\toprule
\textbf{Method} & \textbf{GPU hours}$\downarrow$ & \textbf{Accuracy}$\uparrow$ & \textbf{HSPP}$\downarrow$ \\
\midrule
Vanilla & \textbf{1.30} & 63.18 & 65.51 \\
Reprompt & 3.20 & 58.43 & 64.56 \\
Self-Consistency & 15.00 & 63.74 & 69.13 \\
Long Reasoning & 2.50 & 57.26 & 62.50 \\
Judgment SFT & 7.70 & 61.80 & 54.59 \\
\textbf{\method{}} ($K$=3) & 10.70 & 65.68 & 52.62 \\
\textbf{\method{}} ($K$=5) & 13.30 & \textbf{66.70} & \textbf{50.58} \\
\textbf{\method{}} ($K$=7) & 15.90 & 66.47 & 52.11 \\
\textbf{\method{}} ($K$=9) & 18.50 & \textbf{66.70} & 51.47 \\
\bottomrule
\end{tabular}
\caption{Cost–Performance Tradeoff Analysis. We use Qwen2.5-7B-Instruct as the backbone model.}
\label{table1:cost}
\end{table}

\subsection{Cost--Performance Tradeoff Analysis}
In this section, we further analyze the tradeoff between cost and performance for \method{} to evaluate its scalability.

As shown in Table~\ref{table1:cost}, we conduct experiments on Qwen2.5-7B-Instruct, using total GPU hours as the cost metric and average judgment accuracy and average HSPP as the performance metrics.
\method{} relies on offline multi-agent response sampling during the data construction stage. Therefore, as the number of agents or the number of sampling operations per agent ($K$) increases, the overall data construction cost inevitably rises. Importantly, this cost depends solely on $K$ and the number of agents, and does not increase with input length.
During inference, all baseline methods and Genii variants with different $K$ values require only a single forward pass through the backbone model, maintaining essentially the same computational complexity. Thus, increasing $K$ only improves the quality of preference data in the offline construction stage, thereby enhancing judgment accuracy and fairness, without introducing any extra inference overhead.
Notably, under this comparable total cost, \method{} achieves consistently higher judgment accuracy and better bias mitigation than baseline methods, and can flexibly boost performance by adjusting $K$, demonstrating strong scalability.

\begin{figure}[t]
\subfigure[RewardBench.] { 
\label{fig:rewardbench:task:judge:acc} 
\includegraphics[width=0.48\linewidth]{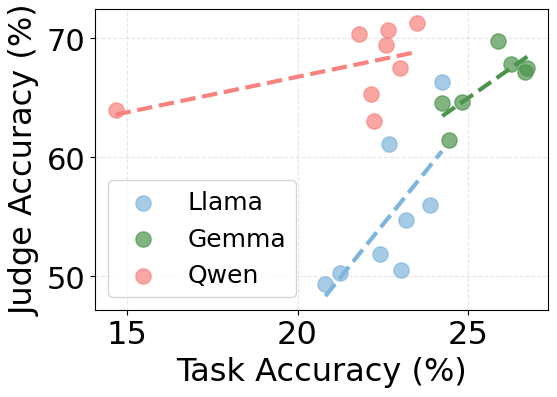}}  
\subfigure[UltraFeedback.] { 
\label{fig:ultrafeedback:task:judge:acc} 
\includegraphics[width=0.48\linewidth]{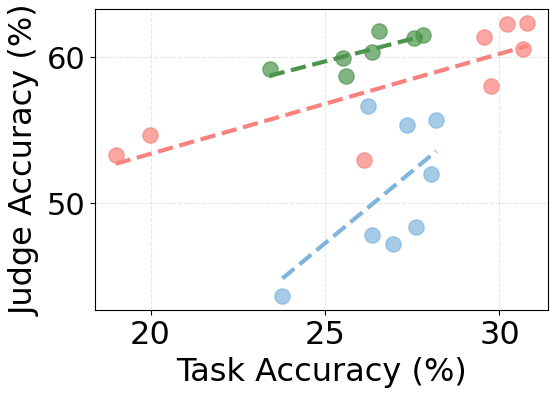}} 
\caption{Correlation between Judgment Accuracy and Task Accuracy. Each circle represents one judgment model, with the color denoting model family.}
\label{fig:task:judge:acc}
\vspace{-1em}
\end{figure}
\begin{figure}[t]
\subfigure[Qwen2.5-7B-Instruct.] { 
\label{fig:qwen:bias:ppl} 
\includegraphics[width=0.48\linewidth]{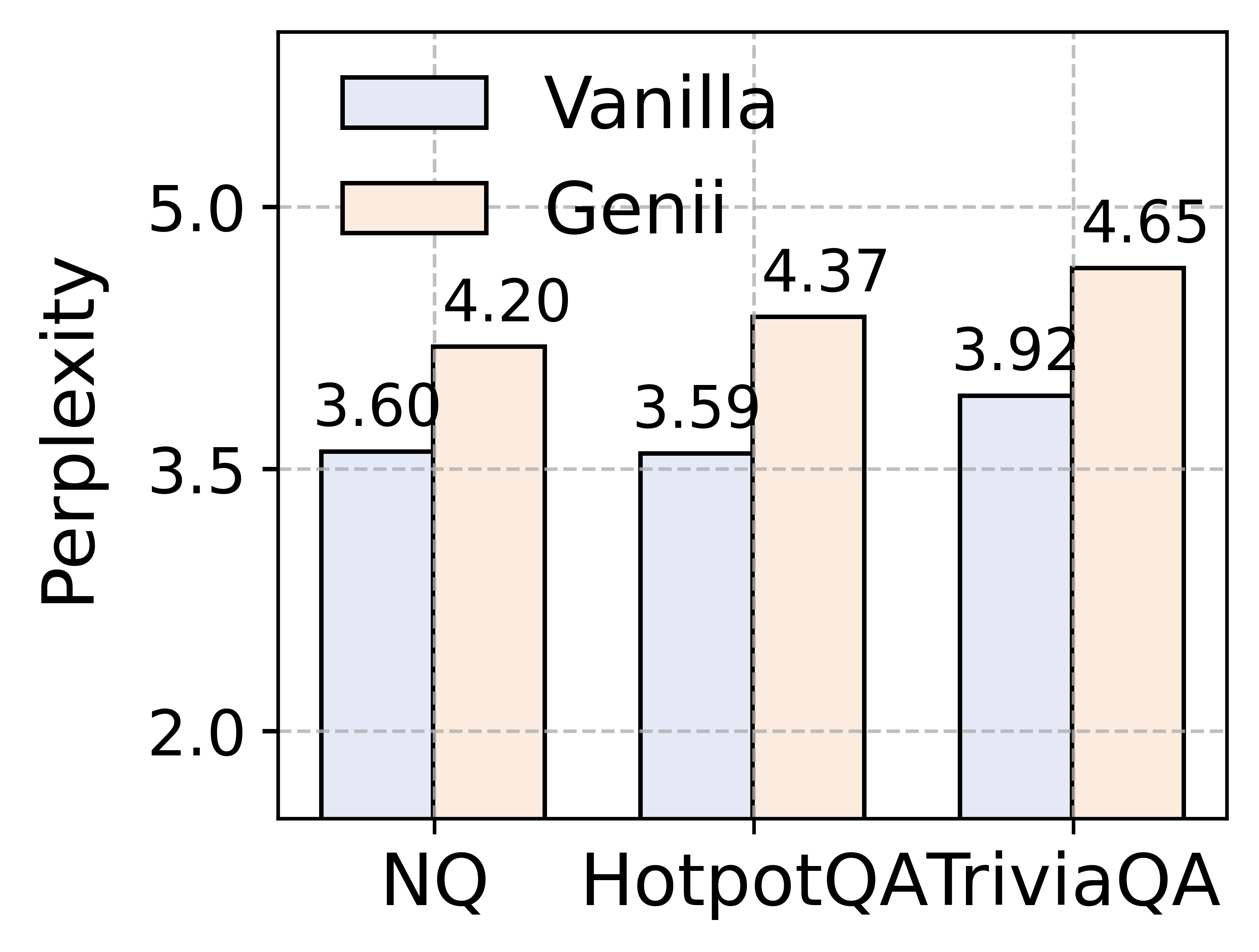}}   
\subfigure[Gemma2-9B-Instruct.] { 
\label{fig:gemma:bias:ppl} 
\includegraphics[width=0.48\linewidth]{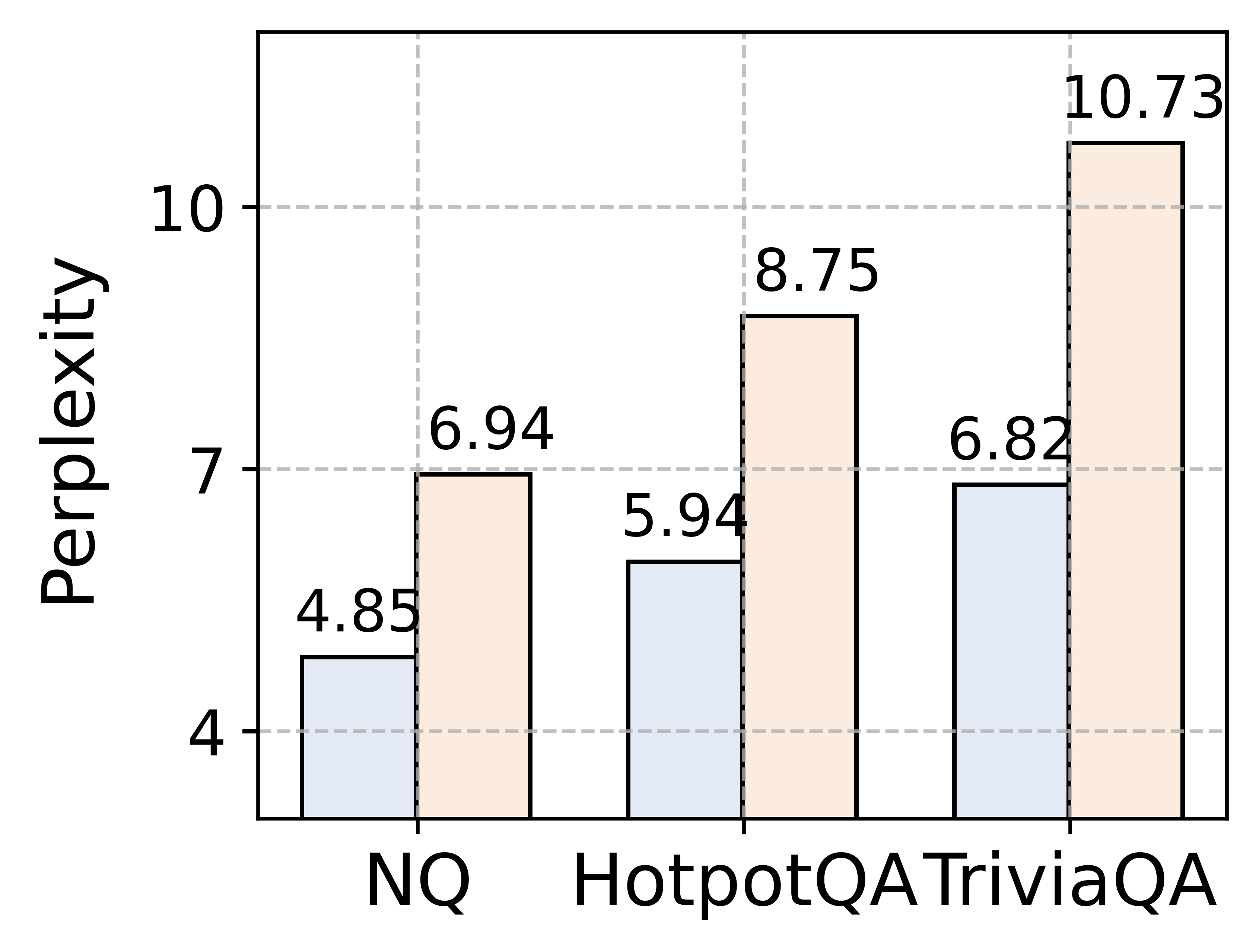}} 
\caption{Perplexity Scores of Vanilla LLMs and \method{} on the Incorrect Answers Generated by Vanilla LLMs. }
\label{fig:bias:ppl}
\vspace{-1em}
\end{figure}

\subsection{Effectiveness of \method{} in Optimizating Judgment Models}
In this section, we first present the relationship between the judgment ability of LLMs and their generation quality. Then, we further explore the effectiveness of \method{} in mitigating preference bias.

\textbf{Judgment–Generation Relationship.} As shown in Figure~\ref{fig:task:judge:acc}, we evaluate LLM-based judgment models introduced in Sec.~\ref{sec:overall} and~\ref{sec:ablation} on multiple benchmarks. On each benchmark, every model is evaluated in two distinct roles: as an evaluator, assessing the model’s ability to identify the better response in pairwise comparisons correctly, and as a generator, measuring whether its generated answer aligns more closely with the ground truth using the ROUGE-L score.
Experimental results demonstrate a robust and broadly positive correlation between judgment accuracy and task accuracy across all datasets and model families. Models that generate higher-quality answers also tend to exhibit greater accuracy in selecting the best response among multiple candidates, highlighting the effectiveness of \method{} in enhancing judgment performance by reinforcing generation capability.
However, despite this overall trend, the strength of the positive correlation varies among models. For instance, Llama3.1-8B-Instruct consistently displays a strong positive correlation between judgment accuracy and task accuracy, whereas Gemma2-9B-Instruct exhibits a markedly weaker relationship.
Meanwhile, Qwen2.5-7B-Instruct shows an unstable linear relationship across benchmarks.

%

\textbf{Mitigating Preference Bias.} We further explore the effectiveness of \method{} in mitigating preference bias. We conduct experiments on the NQ, HotpotQA, and TriviaQA datasets and construct a test dataset by collecting all question–response pairs where the Vanilla model produces an incorrect answer.
We then employ Vanilla LLM and \method{} to reproduce these incorrect answers and calculate the Perplexity (PPL) score.
A lower PPL score indicates that the judgment model is more likely to prefer its own incorrect answers. 

As shown in Figure~\ref{fig:bias:ppl}, the PPL scores of incorrect answers produced by \method{} are significantly higher than those produced by the vanilla LLM across all datasets.
These results show that \method{} exhibits a lower preference bias toward its own incorrect answers. This demonstrates that \method{} can effectively mitigate judgment preference bias by aligning the model's judgments with the collective preference of the group.
Furthermore, Qwen2.5-7B-Instruct, which already exhibits strong judgment performance, experiences only a slight reduction in preference bias after optimization with \method{}. In contrast, the weaker model Gemma2-9B-Instruct achieves a more substantial improvement. This indicates that \method{} provides more pronounced improvements for weaker models compared to stronger ones.

\section{Conclusion}
\label{sec:conclusion}
This paper introduces the \textbf{G}roup-Bas\textbf{e}d Polli\textbf{n}g Opt\textbf{i}m\textbf{i}zation (\method{}) method, an unsupervised multi-agent collaborative optimization framework that encourages multiple LLM-based judgment models to interact with each other to mitigate their own judgment preference bias. 
By interactive client-server polling, \method{} not only enables strong models to guide weaker ones but also leverages feedback from weaker models to optimize stronger ones meaningfully.

\section*{Limitations}
Although \method{} demonstrates convincing performance in enhancing the judgment ability and mitigating the preference bias of LLMs, it has some limitations. First, \method{} employs the strong MiniCPM-Embedding model to compute similarity scores between responses. However, the quality of this embedding-based similarity may constrain the overall performance of \method{}. Further exploration of more fine-grained approaches could enhance its effectiveness. Second, \method{} has demonstrated effectiveness on small models. However, we do not conduct experiments on larger-scale LLMs because of resource constraints. Extending \method{} to larger-scale LLMs will enable the development of more powerful and reliable judgment models.

\section*{Acknowledgments}
This work is supported by the National Natural Science Foundation of China (No. 62576082). This work is also supported by the AI9Stars community.

\bibliography{reference}

\clearpage
\appendix
\section{Appendix}
\label{sec:appendix}
  

\begin{table}[t]
\centering
\small
\begin{tabular}{llr}
\toprule
\textbf{Split} & \textbf{Dataset} & \textbf{Total} \\
\midrule
\multirow{6}{*}{Training} 
  & Evol-Instruct~\shortcite{xu2024wizardlm} & 5,328 \\
  & ASQA~\shortcite{Stelmakh2022ASQA} & 190 \\
  & MARCO QA~\shortcite{bajaj2016marco} & 675 \\
  & Web Questions~\shortcite{Berant2014webquestion} & 754 \\
  & WikiQA~\shortcite{Bajaj2015wikiqa} & 208 \\
  & Yahoo!QA & 837 \\
\midrule
Development
  & UltraFeedback~\shortcite{cui2023ultrafeedback} & 500 \\
\bottomrule
\end{tabular}
\caption{Data Statistics of the Training and Development Datasets.}
\label{table2:traindataset}
\end{table}


\subsection{License}
\label{app:license}
We show the licenses for the datasets that we use. Preferencebench does not report the license of the dataset in the paper or a repository.
PreferenceBench shows its terms of use at website\footnote{\url{https://huggingface.co/datasets/prometheus-eval/Preference-Bench}}.
Yahoo!QA shows its terms of use at website\footnote{\url{https://tensorflow.google.cn/datasets/community_catalog/huggingface/yahoo_answers_qa}}.
We summarize the licenses of the remaining datasets as follows:

All of these licenses and agreements allow their data for academic use: NQ (CC BY-SA 3.0 license); Web Questions, HotpotQA, MTBench, AutoJ (CC BY 4.0 license); MARCO QA, WikiQA, UltraFeedback, and Evol-Instruct (MIT license); TriviaQA, ASQA, and RewardBench (Apache 2.0 license).

\subsection{Additional Experimental Details}
\label{more_details}
In this section, we provide more detailed descriptions of our experiments. We first present the details of the training data used for \method{}. Then, we describe the process of constructing the dataset for assessing the Harmful Self-Preference Propensity (HSPP) of judgment models. Finally, we introduce the implementation details of the experiments.

\textbf{Training Data.}
As shown in Table~\ref{table2:traindataset}, to construct the \method{} training dataset, we
collect instructions from Evol-Instruct~\cite{xu2024wizardlm} and QA pairs from five widely used QA datasets: ASQA~\cite{Stelmakh2022ASQA}, MARCO QA~\cite{bajaj2016marco}, WebQuestions~\cite{Berant2014webquestion}, WikiQA~\cite{Bajaj2015wikiqa}, and YahooQA. Then we further integrate three LLM-based judgment models into a collaborative multi-agent system, including Llama3.1-8B-Instruction~\cite{touvron2023llama}, Gemma2-9B-Instruction~\cite{team2024gemma}, and Qwen2.5-7B-Instruction~\cite{Bai2023qwen}. During the training phase, each agent performs 5 response–request sampling operations.

\textbf{Evaluation Data.} To construct the dataset for assessing the HSPP of each judgment model in the agent system, we first treat each judgment model in turn as the target and pair it with the other two models, resulting in two distinct pairings for each target model.
Within each pair, both models independently generate a response for every question in the evaluation set.
We retain only those query-response pairs where exactly one response is correct and the other is incorrect, with correctness determined by whether a response contains the ground truth.
This process produces two different datasets for each target model, which are then merged into a unified evaluation dataset.
For each sample, we swap the positions of the two candidate responses and evaluate twice to mitigate position bias.

\textbf{Implementation Details.}
In our experiments, we use LoRA~\citep{hu2022lora} for efficient fine-tuning of LLMs. The maximum epochs is set to 3, with a learning rate of $5e-5$ and a warmup ratio of 0.1. For judgment models used in experiments, we use Llama3.1-8B-Instruct~\cite{touvron2023llama}, Gemma2-9B-Instruct~\cite{team2024gemma}, and Qwen2.5-7B-Instruct~\cite{Bai2023qwen} as the backbone models. 
Besides, we use MiniCPM-Embedding\footnote{\url{https://huggingface.co/openbmb/MiniCPM-Embedding}} to compute the similarity scores of responses. 
We perform 10 independent evaluations for the Self-Consistency baseline and aggregate the results via majority voting.
For the Long Reasoning baseline, we employ DeepSeek-R1-Distill-Llama-8B and DeepSeek-R1-Distill-Qwen-7B as the judgment models~\cite{guo2025deepseek}.

\begin{figure}[t]
    \centering
    \subfigure[Vanilla LLM.] { 
        \label{fig:vanilla:tsne} 
        \includegraphics[width=0.45\linewidth]{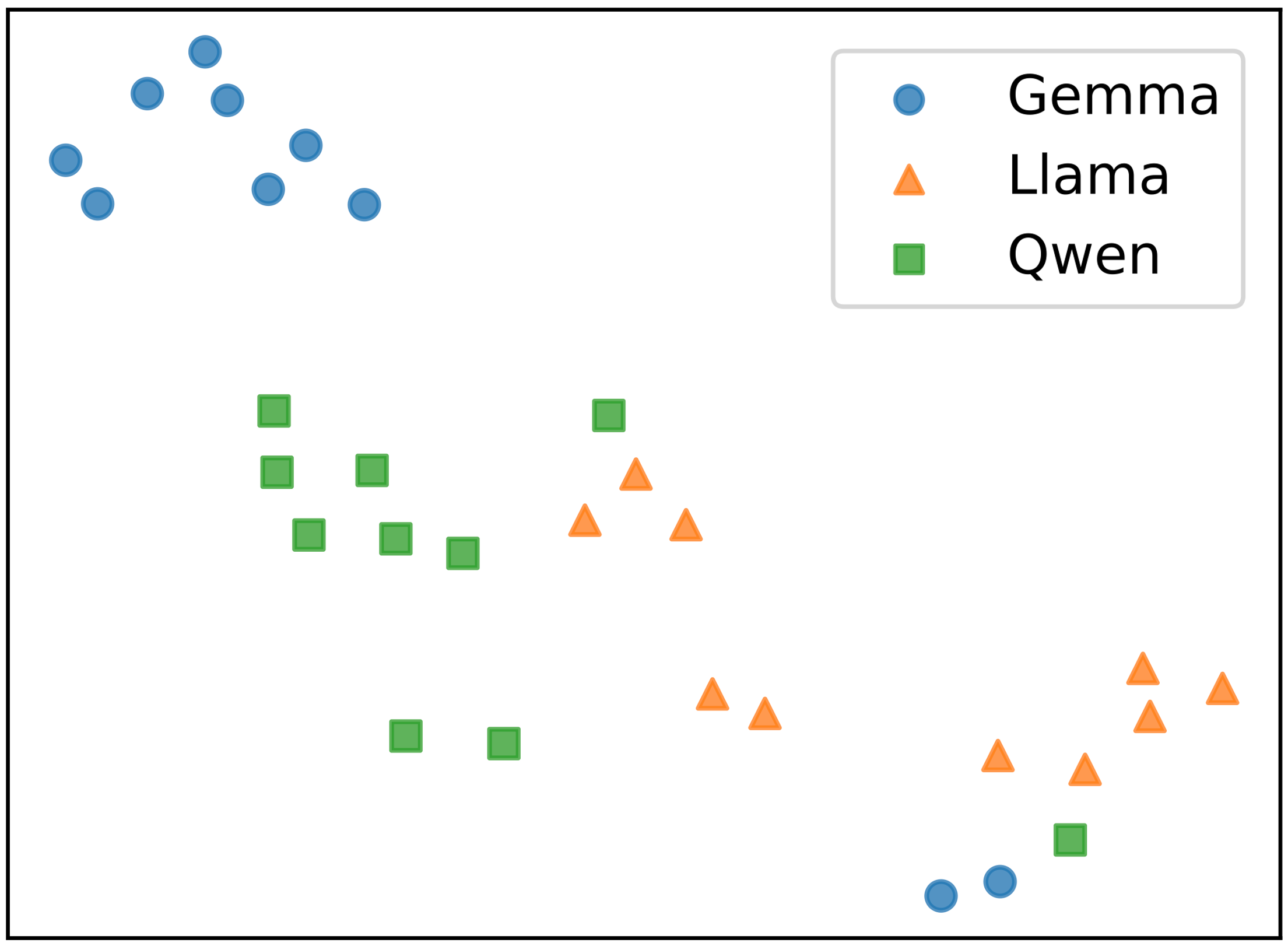}
    }
    \subfigure[\method{}.] { 
        \label{fig:genii:tsne} 
        \includegraphics[width=0.45\linewidth]{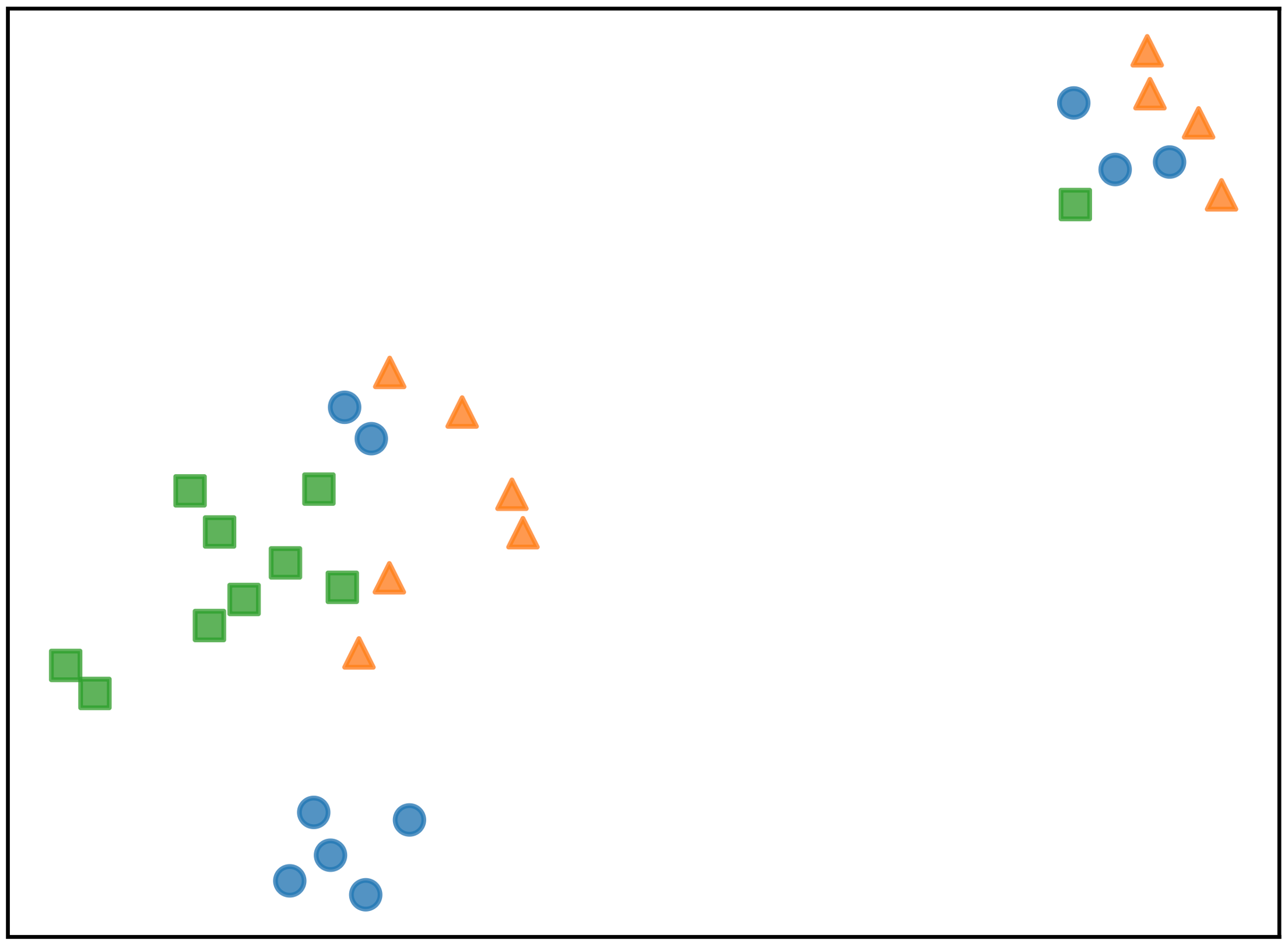}
    }
    \caption{Embedding Visualization of Judgments. We randomly select a sample from the evaluation datasets and then use t-SNE to visualize the embeddings of judgments generated by the Vanilla LLM and \method{}.}
    \label{fig:tsne}
\end{figure}

\subsection{The Judgment Consistency of Vanilla LLMs and \method{}}
\label{app:judgment_consistency}
In this section, we further analyze the consistency of judgments generated by the Vanilla LLM and \method{}. We randomly select a sample from the evaluation datasets. For each model in the multi-agent system, we use both the Vanilla LLM and \method{} to generate 10 independent judgments for this sample. We collect all judgments generated by the Vanilla LLM and all those generated by \method{}, resulting in two separate datasets.
We then employ MiniCPM-embedding to obtain the representations of judgments and visualize them.

As shown in Figure~\ref{fig:tsne}, compared to the embeddings of judgments generated by the Vanilla LLM, those produced by \method{} are more tightly clustered, indicating that \method{} achieves higher consistency in judgments of different LLMs. 
This further demonstrates the effectiveness of \method{} in aligning each agent in the system with the collective preference of the group to yield more accurate judgments.

\begin{figure}[t]
\centering
\includegraphics[width=0.9\linewidth]{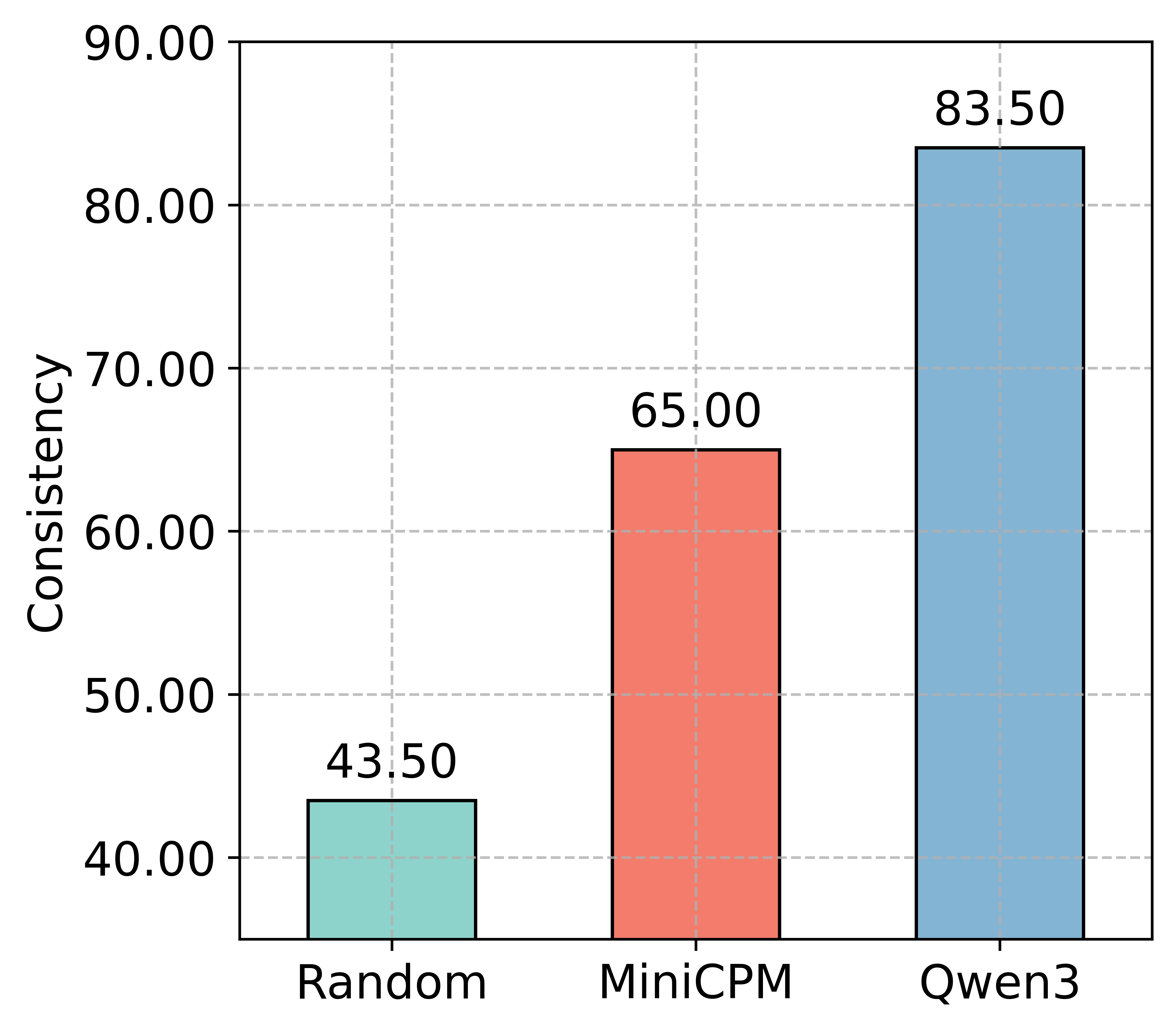}
\caption{Agreement between the Consistency Score and Human Judgments.}
\label{fig:judge:agreement}
\vspace{-1em}
\end{figure}

\subsection{Agreement between the Consistency Score and Human Judgments}
\label{app:judgment_aggrement}
In this subsection, we further evaluate the effectiveness and robustness of the embedding-based consistency approach through an additional experiment. We randomly sample 200 instances from the Evol-Instruct dataset, with each instance containing six candidate responses generated by different agent models. For each instance, we compare three strategies for selecting the best response: selecting at random, using the embedding-based consistency score with MiniCPM-Embedding, and using the embedding-based consistency score with Qwen3-Embedding-8B. Using human annotations as the gold standard, we compute the accuracy of each strategy by assessing whether the selected response matches the human-annotated gold answer.

As shown in Figure~\ref{fig:judge:agreement}, the embedding-based strategies achieve substantially higher accuracy than random selection, demonstrating over 60\% agreement with human judgments. These results highlight the effectiveness and robustness of the embedding-based consistency score for unsupervised preference data construction.

\begin{table*}[t]
\centering
\small
\begin{tabular}{lcccccc}
\toprule
\multirow{2}{*}{\textbf{Method}}
& \multicolumn{2}{c}{\textbf{PreferenceBench}} 
& \multicolumn{2}{c}{\textbf{RewardBench}}
& \multicolumn{2}{c}{\textbf{UltraFeedback}} \\
\cmidrule(lr){2-3} \cmidrule(lr){4-5} \cmidrule(lr){6-7}
& \textbf{position1} & \textbf{position2}
& \textbf{position1} & \textbf{position2}
& \textbf{position1} & \textbf{position2} \\
\midrule
\multicolumn{7}{l}{\textit{Llama3.1-8B-Instruct}} \\
\midrule
Vanilla & 82.00 & 55.00 & 74.00 & 30.00 & 56.78 & 24.75 \\
\textbf{\method{}} & 64.58 & 79.08 & 58.12 & 57.22 & 48.90 & 49.23 \\
\midrule
\multicolumn{7}{l}{\textit{Gemma2-9B-Instruct}} \\
\midrule
Vanilla & 67.84 & 84.00 & 70.05 & 68.34 & 55.05 & 58.50 \\
\textbf{\method{}} & 81.50 & 85.00 & 74.00 & 66.00 & 62.50 & 57.50 \\
\midrule
\multicolumn{7}{l}{\textit{Qwen2.5-7B-Instruct}} \\
\midrule
Vanilla & 85.00 & 76.50 & 78.00 & 59.80 & 69.00 & 54.82 \\
\textbf{\method{}} & 86.50 & 91.00 & 83.00 & 65.50 & 68.00 & 57.00 \\
\bottomrule
\end{tabular}
\caption{Judgment Accuracy of Different LLMs When the Human-Annotated Best Response is Placed at Different Positions in the Response Sequence. ``position-$i$'' indicates that the gold response occupies the $i$-th position, reflecting each model’s robustness to position bias in evaluation.}
\label{tab:position_bias}
\end{table*}

\subsection{Mitigating Position Bias in Evaluation}
\label{position_bias}
This section shows the robustness of \method{} to position bias. We randomly sample 200 queries from each of the three datasets--PreferenceBench, RewardBench, and UltraFeedback.

For each query, we construct evaluation instances in which the human-annotated best response is deliberately placed at different positions within the response set.
Specifically, the best response is positioned in the 1st and 2nd positions among the two main choices in the prompt, while the remaining positions are filled with other candidate responses.
This procedure yields two distinct test scenarios where the gold response occupies different positions, allowing us to directly assess the model’s robustness to positional bias.
By comparing the judgment accuracy of the Vanilla LLM and \method{} across these positional configurations, we evaluate their capability to mitigate position-induced bias.
A model that maintains consistently high accuracy across all positional conditions is considered less susceptible to positional bias during evaluation.

As shown in Table~\ref{tab:position_bias}, \method{} consistently outperforms the Vanilla LLM across all datasets, demonstrating its effectiveness in reducing position bias and improving evaluation reliability.

\subsection{Prompt Templates Used in Experiments}
\label{prompt}
In this section, we present the prompt templates used in our experiments.

\begin{figure*}[h]
\centering
\includegraphics[width=0.8\linewidth]{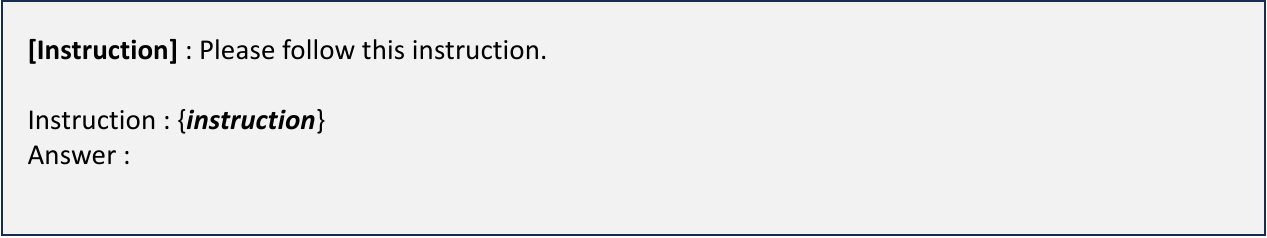}
\caption{The Prompt Templates Used for Training \method{}.} 
\label{fig:train_prompt}
\end{figure*}

\begin{figure*}[h]
\centering
\includegraphics[width=0.8\linewidth]{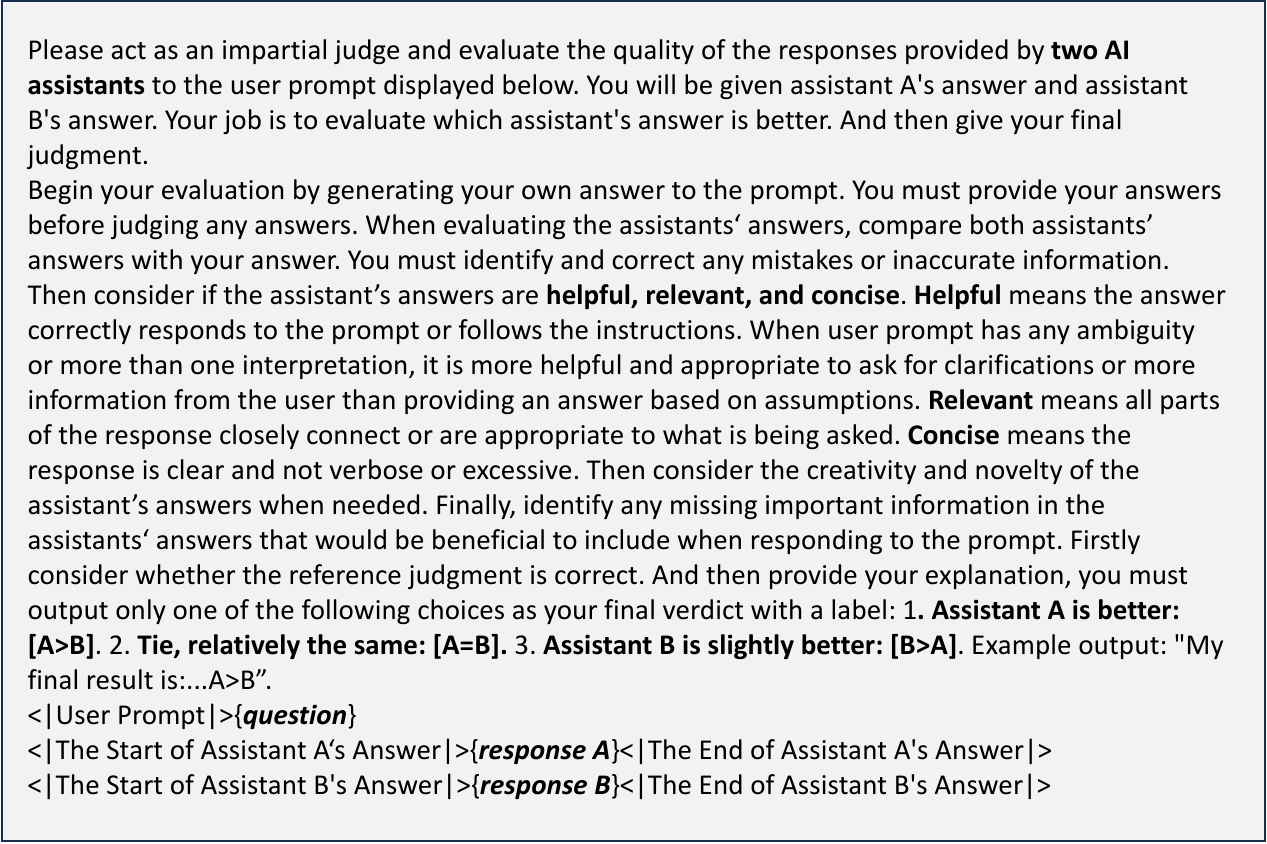}
\caption{The Prompt Templates Used for Assessing Accuracy.} 
\label{fig:judgment_acc}
\end{figure*}

\begin{figure*}[h]
\centering
\includegraphics[width=0.8\linewidth]{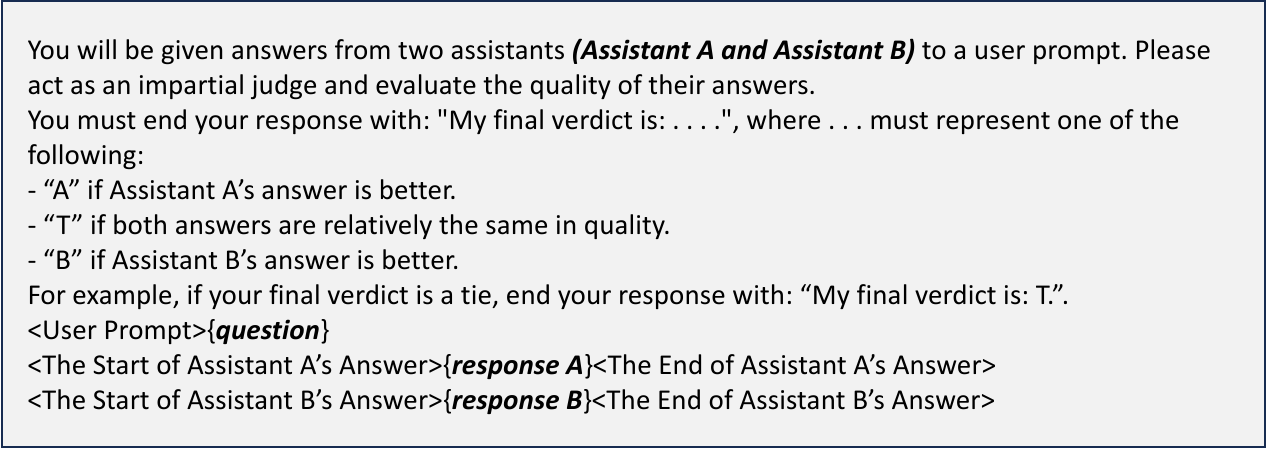}
\caption{The Prompt Templates Used for Assessing Harmful Self-Preference Bias.} 
\label{fig:judgment_preference_bias}
\end{figure*}

\begin{figure*}[!t]
\centering
\includegraphics[width=0.8\linewidth]{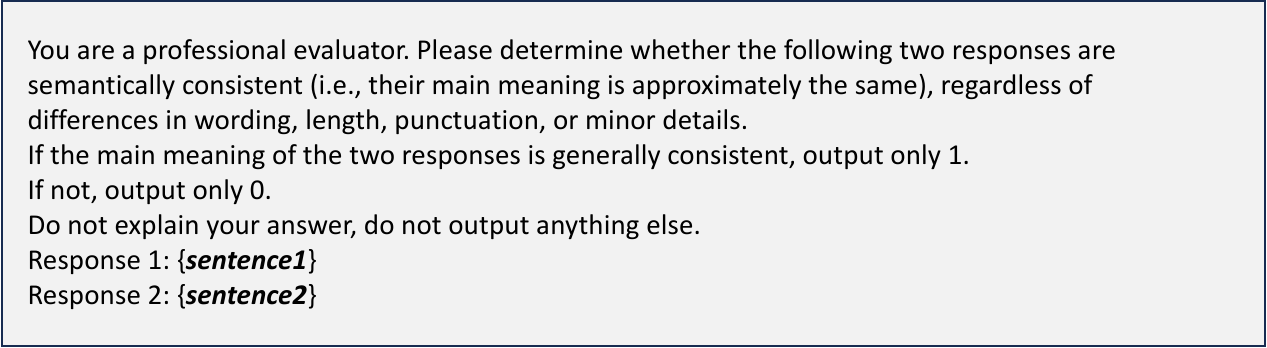}
\caption{The Prompt Templates Used for GLM-4.5 to Evaluate the Quality of the Responses.} 
\label{fig:agreement_prompt}
\end{figure*}

First, we present the prompt designed for polling sampling and \method{} training, as shown in Figure~\ref{fig:train_prompt}. 
Next, as illustrated in Figures~\ref{fig:judgment_acc} and ~\ref{fig:judgment_preference_bias}, we introduce the prompts used for evaluating judgment accuracy and harmful self-preference propensity of judgment models. We follow~\cite{zheng2023judging,chen2025llm} to design these prompt templates, which facilitate the generation of more accurate evaluation results.
Finally, we show the prompt template used in Appendix~\ref{app:judgment_aggrement} to evaluate the quality of the responses chosen by two different selection strategies, as shown in Figure~\ref{fig:agreement_prompt}.


\subsection{Case Study}
\label{case}
In Table~\ref{tab:case1}, we present two cases to show the effectiveness of the \method{} method. 

In the first case, we compare the judgment accuracy of different judgment models.
For a given query, along with two candidate responses of differing quality, each judgment model selects the better answer under its own criteria. Vanilla LLM favors answers that use familiar language and straightforward puns, leading it to prefer the more conventional Response A over the higher-quality Response B.
The Long Reasoning model places too much emphasis on explicit alignment with user intent and character traits, overlooking the objective creativity and engagement of Response B.
Judgment SFT, from the perspective of narrative flow and tight integration of humor, tends to choose Response A that feels more coherent and polished.
In contrast, only \method{} truly satisfies the requirements of the question and correctly selects the higher-quality Response B.
This demonstrates the effectiveness of \method{} in providing accurate evaluations.

In the second case, we compare how different judgment models exhibit preference bias when evaluating their own incorrect responses and correct responses produced by other models. 
Vanilla LLM prioritizes explanation structure and comprehensiveness, which leads it to overlook factual accuracy (``Delaware'') in Response B and incorrectly prefer it.
Similarly, Long Reasoning prioritizes answers that appear more informative, so it also prefers its own incorrect response.
Judgment SFT pays more attention to the presence of detailed step-by-step reasoning, which may result in a bias toward responses like B that are less accurate.
In contrast, \method{} pays attention to the objective factual content and context, and correctly prefers the correct Response A (``Pittsburgh''), showing its effectiveness in reducing judgment preference bias.

\begin{table*}[!t]
\centering
\renewcommand\arraystretch{1.5}
\small
\resizebox{\textwidth}{!}{
\begin{tabular}{l}
\hline
\multicolumn{1}{l}{\textit{\textbf{Case \#1 \qquad Judgment Accuracy}}} \\ \hline
\textbf{Question:} Can you help me write a dialogue that reflects their personalities and the humor
style they represent? \\
To what extent does the model successfully integrate humor, irony, or cleverness in appropriate responses? \qquad\qquad \\
\textbf{Response A:} Jake: ``Mia, you remember that confusing case about the disputed inheritance...?'' Mia: `` Oh, the \\ one
where the rich aunt left her fortune to her cats?''
Jake: ``Yes... Now we've got two more \\
beneficiaries claiming their rights.''
\\Mia: ``Two more? Sounds like a... furry situation.'' 
Jake: ``Indeed, a very hairy problem...'' ... \\

\textbf{Response B:} ... Mia: ``Oh, you mean that purr-ticularly hairy case? Sure sounds like a catastrophe.''
Jake: ``It \\ does...
But there's a talking parrot involved now.''
Mia: ``A parrot? Sounds like an animal farm novel... It's \\ getting
beak-ome serious!''
Jake: ``If it doesn’t ruffle too many feathers... ''
Mia: ``You're in a squawkward \\ situation...
but the tweet shall set you free!'' \\
\textbf{Chosen:} Response \sethlcolor{lightgreen}\hl{B}
\\ \cdashline{1-1}
\textbf{Vanilla LLM:} ... Both answers are helpful, relevant, and concise, but Assistant A's dialogue might be \\
\sethlcolor{lightpurple}\hl{slightly more relatable} due to its direct use of puns and wordplay that are more \\ common in everyday language.
My final result: \sethlcolor{lightpurple}\hl{A>B}.\\
\textbf{Long Reasoning:} ... Overall, both answers are good, but Assistant A's response \sethlcolor{lightpurple}\hl{better aligns with the user}'s \\
specifications 
and provides a \sethlcolor{lightpurple}\hl{more accurate representation} of Mia's personality and the required humor.\\
</think>My final result: \sethlcolor{lightpurple}\hl{A>B}. \\
\textbf{SFT Judgment:} ... Overall, A’s adherence to the characters’ established traits, tighter integration of humor, \\
and natural narrative flow make
it a \sethlcolor{lightpurple}\hl{stronger response compared to B}’s disjointed approach.\\
My final result: \sethlcolor{lightpurple}\hl{A>B}.\\
\textbf{\method{}:} ... Both answers are helpful and relevant, but Assistant B's answer ...  \sethlcolor{lightpinkred}\hl{more engaging and creative}. \\
This slight edge 
in creativity and novelty makes Assistant B's answer \sethlcolor{lightpinkred}\hl{stand out}. My final result: \sethlcolor{lightpinkred}\hl{B>A}.\\
\hline

\multicolumn{1}{l}{\textit{\textbf{Case \#2 \qquad Judgment Preference Bias}}} \\ \hline
\textbf{Question:} Where does the ``This Is Us'' family live? \qquad\qquad \textbf{Answer:} \sethlcolor{lightgreen}\hl{Pittsburgh.}  \\
\textbf{Response A (Gemma):} The show \"This Is Us\" primarily takes place in **\sethlcolor{lightgreen}\hl{Pittsburgh}, Pennsylvania**. 
Here's \\
a breakdown:* **Pearson  
family home in the present day:** The majority of the present-day scenes are set \\ in various 
locations around suburban Pittsburgh ... \\
\textbf{Response B (Qwen):} The Pearson family from the TV show \"This Is Us\" primarily
lives in \sethlcolor{lightred}\hl{Delaware}, \\specifically in the city of Garden City, 
Delaware. While the precise address is not disclosed on the show...
\\ \cdashline{1-1}
\textbf{Vanilla LLM (Qwen):} Assistant A provides a detailed breakdown of the locations, ... , However, Assistant \\ A's answer
is \sethlcolor{lightpurple}\hl{not entirely accurate}, ... Assistant B provides a \sethlcolor{lightpurple}\hl{comprehensive explanation}, ... My final \\
verdict is: \sethlcolor{lightpurple}\hl{B}.\\
\textbf{Long Reasoning (Qwen):} Assistant A gives a \sethlcolor{lightpurple}\hl{straightforward} answer focusing on Pittsburgh... Assistant \\
B provides a more thorough answer with details and the show's style, making it \sethlcolor{lightpurple}\hl{more informative} ... \\
My final verdict is: \sethlcolor{lightpurple}\hl{B}. \\
\textbf{SFT Judgment (Qwen):} Assistant A provides a detailed breakdown, ... However, the information seems \\
to \sethlcolor{lightpurple}\hl{contradict} the
show's actual setting, ... , Assistant B, on the other hand, \sethlcolor{lightpurple}\hl{accurately} ...
My final verdict is: \sethlcolor{lightpurple}\hl{B}.\\
\textbf{\method{} (Qwen):} Assistant A offers a broader perspective that includes more specific details, ... 
while Assistant \\ 
B gives a \sethlcolor{lightpinkred}\hl{more focused view on the main setting} and some additional context.My final verdict is: \sethlcolor{lightpinkred}\hl{A}.\\
\hline
\end{tabular}
}
\caption{\label{tab:case1}Case Studies. If the phrase aligns with the facts, it is highlighted in \sethlcolor{lightgreen}\hl{Blue}, while inconsistencies are highlighted in \sethlcolor{lightred}\hl{Orange}.  
Correct evaluations are marked in \sethlcolor{lightpinkred}\hl{Red}, whereas incorrect results are highlighted in \sethlcolor{lightpurple}\hl{Purple}.}
\end{table*}

\end{document}